%% file: main_arxiv.tex
\documentclass{uber}

\definecolor{gtgreen}{HTML}{39FF14}
\definecolor{predorange}{HTML}{FFB300}

\renewcommand\contributionformat[2][]{{\small $#1$\,#2}}
 
\title{TraVEL: Trajectory-Guided Video Embedding Learning for Driving-Video Retrieval}

\author{Yi-Chung Chen$^{1,2}$}
\author{Philip Jacobson$^{1\dagger}$}
\author{Tom Lampo$^{1}$}
\author{Yiren Lu$^{1}$}
\author{Jin Yao$^{1}$}
\author{David I. Inouye$^{2}$} 
\author{Jing Gao$^{2}$}
\author{Danhua Guo$^{1}$}
\author{Burhan Yaman$^{1\ddagger}$}
\affiliation[1]{Uber AV Labs}
\affiliation[2]{Purdue University}

\contribution[\dagger]{Corresponding author}\contribution[\ddagger]{Project Lead}

\date{\today}
\renewcommand{\uberlogo}{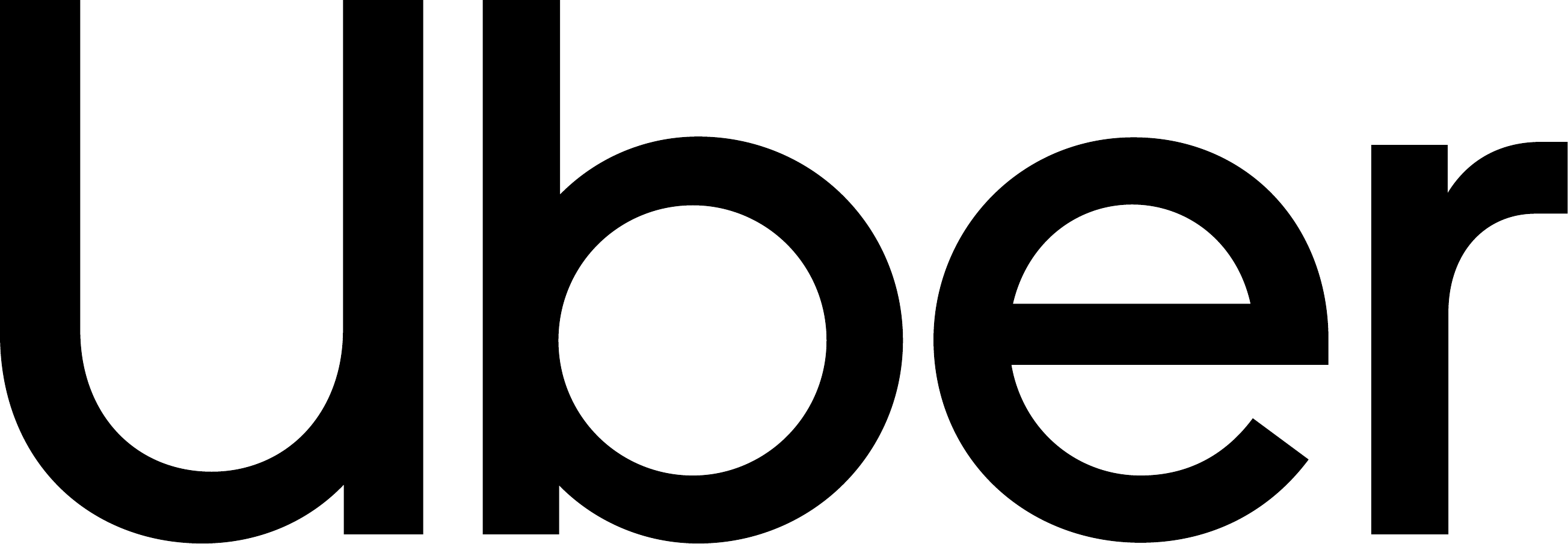}
\renewcommand{\uberlogosecond}{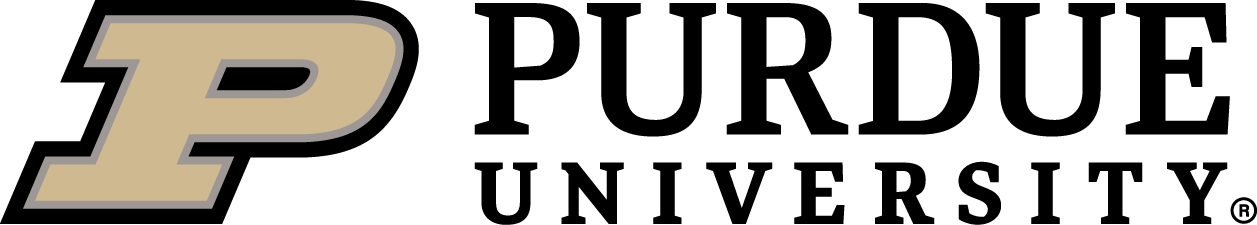}

\abstract{%
Efficiently retrieving relevant clips from large-scale driving logs is essential
for data curation, model development, and safety analysis. Structured and
rule-based retrieval systems can explicitly target driving events, but typically
require expert-defined rules, auxiliary data, and multi-stage perception
pipelines. Multimodal embedding models offer a simpler and more efficient
alternative by representing each video with a single searchable vector. However,
general-purpose models often rely on shortcuts from static scene context and
struggle to distinguish motion-centric events, such as turning left versus right
or accelerating versus decelerating. In this work, we study how to adapt a
general-purpose multimodal embedding model to driving-video retrieval. We first
fine-tune Qwen3-VL-Embedding on paired clips and reasoning traces from
nuReasoning using an InfoNCE objective. While this stage substantially improves
overall retrieval, caption supervision alone remains insufficient for
fine-grained motion understanding. We therefore introduce TraVEL
(Trajectory-Guided Video Embedding Learning), a motion-aware fine-tuning
framework that uses ego-trajectory similarity as a reward within Group Relative
Policy Optimization. Trajectories serve only as privileged training supervision;
retrieval still operates on single-vector video embeddings without ego poses,
expert rules, or auxiliary perception outputs. We further construct a
driving-video retrieval benchmark from nuReasoning. Experiments show that TraVEL
improves motion-centric retrieval across model scales: relative to SFT, it raises
longitudinal and lateral mAP by 9.8 and 4.7 points at 2B, with corresponding gains of 7.2 and 1.5 points at
8B. TraVEL thus combines physically grounded supervision with efficient
embedding-based search.%
}

\begin{document}
\maketitle

\input{sec/1_intro}
\input{sec/2_related_work}
\input{sec/3_preliminaries}
\input{sec/4_methodology}
\input{sec/5_experiments}
\input{sec/6_conclusion}

\bibliographystyle{plainnat}
\bibliography{main}

\end{document}

%% file: sec/1_intro.tex
\section{Introduction}
\label{sec:intro}

Large-scale autonomous-driving fleets continuously collect vast amounts of video, yet safety-critical interactions and long-tail events constitute only a small fraction of these logs. Retrieving scenarios of interest is therefore a critical component of autonomous-driving data engines: retrieved examples can be curated for targeted annotation and model improvement~\cite{liang2024aide}, selected for simulation and safety validation~\cite{davidson2026refav,ding2024realgen}, or used as precedents for counterfactual assessment of candidate driving actions~\cite{patrikar2025negative}. Such targeted data curation can support the development and evaluation of downstream driving systems~\cite{lu2026splat2bev,yao2026vlga}. The value of these downstream workflows depends on whether operationally meaningful events can be reliably surfaced from large-scale driving data.

In practice, a common and dependable way to retrieve well-specified driving events is to use structured or rule-based systems that translate expert knowledge into functions over detected events and motion cues~\cite{yao2026driving}. Such systems provide explicit control over the target scenario, but the relevant event definitions and decision rules must be designed in advance and calibrated to the domain. Their execution also commonly depends on auxiliary data or multi-stage perception outputs, such as object detections, tracks, map elements, and ego-motion signals. Extending them to new event vocabularies, datasets, or operating conditions can therefore incur substantial engineering cost.

In contrast, multimodal embedding models offer a direct and efficient retrieval interface. They map text and video into a shared representation space, allowing every clip to be precomputed as a single vector and searched through fast similarity matching without a structured event detector or reasoning pipeline at retrieval time. Beginning with CLIP~\cite{radford2021learning}, recent methods have adapted increasingly capable vision-language models into general-purpose embedding models; VLM2Vec, VLM2Vec-V2~\cite{jiang2024vlm2vec,meng2025vlm2vecv2}, and Qwen3-VL-Embedding~\cite{li2026qwen3vlembedding} achieve strong performance across a broad range of multimodal retrieval tasks. However, this simplicity does not guarantee sensitivity to the distinctions that matter in driving. Video--text models often exploit shortcuts based on objects or static scene context and struggle with counterfactual descriptions that differ primarily in their actions~\cite{ma2024rethinking}. Consequently, operationally opposite events, such as turning left versus right or accelerating versus decelerating, can remain difficult to distinguish when they occur in visually similar scenes.

Motivated by this tension between efficient retrieval and motion sensitivity, we ask: \emph{Can a general-purpose multimodal embedding model be adapted into an effective driving-video retriever?} We first evaluate Qwen3-VL-Embedding on driving videos and observe a substantial domain gap: zero-shot retrieval varies widely across actions and is particularly weak for directional turns, lane changes, and subtle within-lane motion. We then fine-tune the model on video clips paired with reasoning-trace captions from the nuReasoning dataset~\cite{huang2026nureasoning} using the InfoNCE objective~\cite{oord2018representation}. Although this supervised fine-tuning stage substantially improves overall retrieval, several motion-centric queries, including acceleration and deceleration, remain challenging even when the corresponding terms appear in the training captions. Caption supervision alone therefore does not provide a sufficiently dense signal for learning fine-grained vehicle motion.

\begin{figure*}[t]
    \centering
    \includegraphics[width=\textwidth]{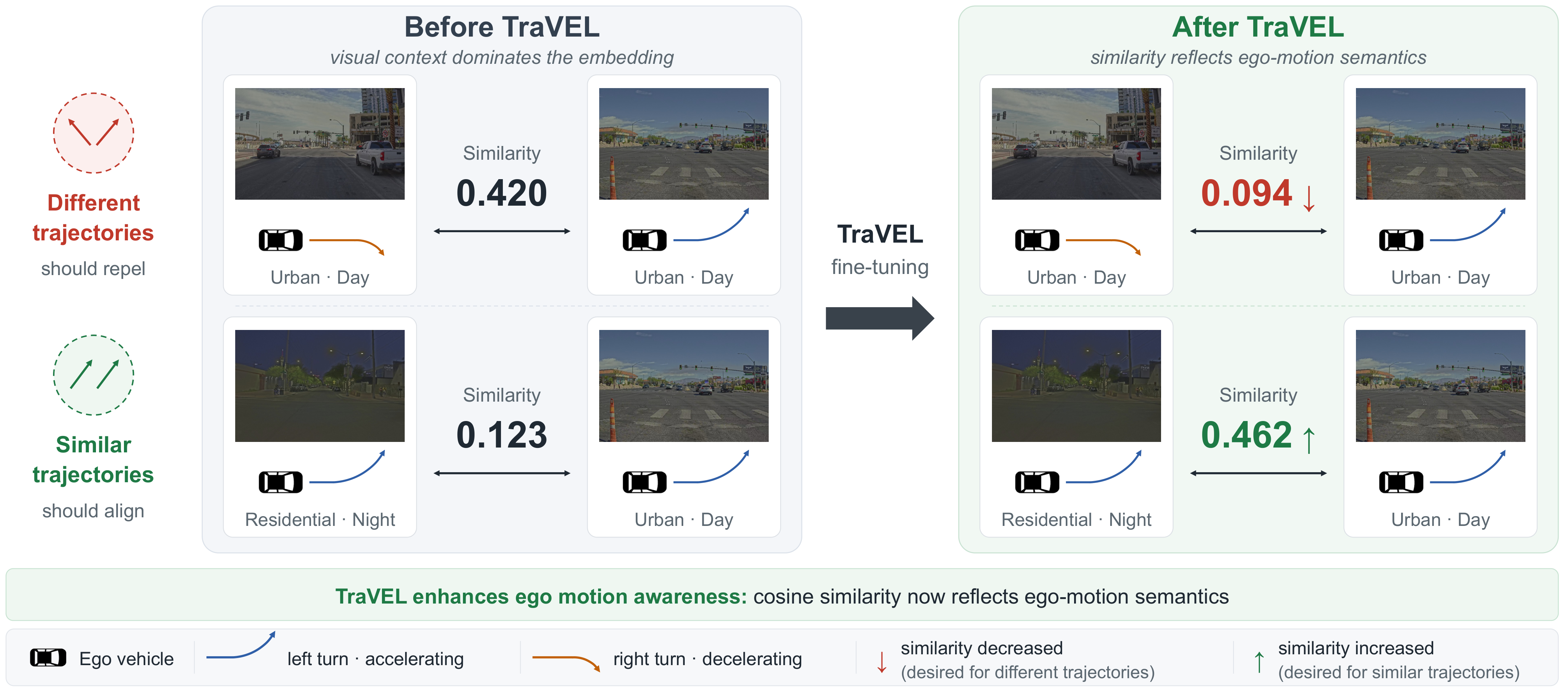}
    \caption{Conceptual effect of TraVEL fine-tuning on the video embedding space. Before trajectory-aware fine-tuning (left), two daytime clips receive relatively high cosine similarity ($0.420$) despite exhibiting different ego motion: an accelerating left turn versus a decelerating right turn. Meanwhile, daytime and nighttime clips that share the same accelerating left-turn trajectory have low similarity ($0.123$). TraVEL reverses this context bias (right), reducing the similarity of the motion-incompatible pair to $0.094$ and increasing the similarity of the matched-trajectory pair to $0.462$. Thus, the learned geometry becomes more sensitive to vehicle motion and less dominated by visual context.}
    \label{fig:travel_embedding_teaser}
\end{figure*}

To address this limitation, we introduce TraVEL (Trajectory-Guided Video Embedding Learning), a motion-aware fine-tuning framework that uses ego-trajectory similarity as a privileged training signal. Ego trajectories are commonly available from onboard localization and inertial sensors; when they are not, they can also be estimated directly from camera observations using visual geometry models such as DVGT~\cite{zuo2026dvgt}. TraVEL incorporates this signal through Group Relative Policy Optimization (GRPO)~\cite{guo2025deepseek}, encouraging the video embedding space to reflect the structure of vehicle motion. As illustrated in Fig.~\ref{fig:travel_embedding_teaser}, TraVEL increases similarity between clips that share the same ego trajectory despite different visual contexts, while decreasing similarity between visually similar clips with incompatible motion. After training, each clip is still indexed and retrieved using a single embedding; ego poses, expert-defined rules, and auxiliary perception outputs are not required at retrieval time. TraVEL thus combines physically grounded motion supervision with the simplicity and efficiency of embedding-based search, yielding particularly clear gains on motion-centric queries.

Our contributions are threefold:
\begin{itemize}
    \item We present TraVEL, a simple and effective framework for adapting general-purpose multimodal embedding models to driving-video retrieval.
    \item We introduce a trajectory-derived reward and a GRPO-based fine-tuning stage that make video embeddings more sensitive to ego motion and improve downstream retrieval performance.
    \item We introduce a driving-video retrieval benchmark derived from nuReasoning and demonstrate the potential of embedding-based retrieval for motion-centric driving queries.
\end{itemize}

%% file: sec/2_related_work.tex
\section{Related Work}
\label{sec:related_work}

\paragraph{General-purpose multimodal embeddings.}
Contrastive vision--language pretraining established shared embedding spaces as an effective interface for large-scale cross-modal retrieval. CLIP~\cite{radford2021learning} learns such a space from image--text pairs, while SigLIP~\cite{zhai2023siglip} replaces softmax normalization over in-batch pairs with independent sigmoid losses. More recent work repurposes pretrained vision-language models as broadly applicable embedding models. VLM2Vec and VLM2Vec-V2~\cite{jiang2024vlm2vec,meng2025vlm2vecv2} unify heterogeneous multimodal tasks under an embedding formulation and extend evaluation from images to videos and visual documents; Qwen3-VL-Embedding~\cite{li2026qwen3vlembedding} further couples a general-purpose embedding model with a reranker. Alternative objectives and interaction mechanisms improve the expressiveness of these representations: VL-JEPA~\cite{chen2025vljepa} predicts continuous language embeddings and supports text-to-video retrieval directly in its latent space, whereas MetaEmbed~\cite{xiao2025metaembed} exposes a compact set of vectors for flexible late interaction at test time. These models demonstrate strong general semantic retrieval, but their ability to distinguish visually similar driving clips by fine-grained ego motion remains underexplored. We retain the efficiency of a single-vector index and adapt its geometry using physical motion available during training.

\paragraph{Fine-grained video--text retrieval.}
Video retrieval methods extend image--text contrastive learning with temporal modeling and video-specific supervision. VideoCLIP~\cite{xu2021videoclip} jointly pretrains video and text transformers, while CLIP4Clip~\cite{luo2021clip4clip} transfers CLIP representations to video retrieval with temporal aggregation. More recently, VeRVE~\cite{halbe2026verve} uses a shared multimodal language-model backbone to support corpus-, moment-, and composed-query retrieval through unified embeddings. A complementary line augments the visual representation with generated language. Cap4Video~\cite{wu2023cap4video} incorporates auxiliary captions into training, feature interaction, and scoring; Narrating the Video~\cite{hur2025narvid} aggregates frame-level captions while filtering irrelevant or incorrect narration; and ELIOT~\cite{liu2025eliot} combines off-the-shelf captioners, language models, and text retrieval for zero-shot search. X-CoT~\cite{pulakurthi2025xcot} instead replaces similarity-only ranking with language-model chain-of-thought comparisons to produce interpretable rankings.

Despite this progress, fine-grained temporal distinctions remain difficult. Counterfactually augmented evaluation reveals that video--text models often rely on objects and scene context rather than the queried action~\cite{ma2024rethinking}. VideoComp~\cite{kim2025videocomp} addresses a related limitation with temporally disrupted negatives and a preference objective for compositional alignment in multi-event videos. Our work shares the goal of motion-sensitive retrieval but takes a different route: rather than adding generated captions, inference-time reasoning, or discrete temporal perturbations, we use continuous ego trajectories to shape the video embedding space during training. Retrieval itself remains a single cosine-similarity search.

\paragraph{Driving-scenario retrieval and mining.}
Driving-log search introduces domain-specific requirements because relevant events are often defined by multi-agent motion and planning context. RefAV~\cite{davidson2026refav} formulates planning-centric scenario mining as detecting and spatiotemporally localizing natural-language descriptions of multi-agent interactions, and evaluates referential multi-object tracking systems on Argoverse~2 logs. CARIM~\cite{ki2025carim} generates compressed scene captions and converts driving-video retrieval into inclusive text-to-text matching so that all requested conditions are satisfied. For complex dynamic events, STRIVE-D~\cite{yao2026driving} calibrates structured query rules from weakly labeled in-domain videos and combines them with vision-language and keyword retrieval signals. These approaches provide expressive control through tracking outputs, intermediate captions, or calibrated rules. In contrast, TraVEL requires no structured event library or auxiliary reasoning pipeline at retrieval time: ego poses act only as privileged training supervision, after which every video is indexed by one embedding and retrieved directly from natural-language queries.

%% file: sec/3_preliminaries.tex
\section{Preliminaries}
\label{sec:preliminaries}

\subsection{Multimodal Embedding-Based Retrieval}
\label{sec:preliminaries:retrieval}

Let $\mathcal{D}=\{(q_i,x_i)\}_{i=1}^{N}$ denote paired natural-language descriptions $q_i$ and videos $x_i$. Multimodal embedding models map variable-length text and visual inputs into a shared fixed-dimensional space. Models such as Qwen3-VL-Embedding~\cite{li2026qwen3vlembedding} tokenize the input content together with a dedicated pooling token; after contextualization by the transformer, the pooling token's final hidden state summarizes the full input and is normalized to form a single embedding vector.

At retrieval time, a text query $q$ ranks a video corpus $\mathcal{X}=\{x_j\}_{j=1}^{M}$. Let $f_\theta$ denote the embedding model, whose outputs are unit-normalized. Each candidate is scored by cosine similarity,
\begin{equation}
    s_\theta(q,x_j)
    = \left\langle f_\theta(q),f_\theta(x_j)\right\rangle.
    \label{eq:retrieval_similarity}
\end{equation}
The candidates are sorted in descending order of $s_\theta(q,x_j)$. Because every corpus video is represented by one precomputed vector, ranking reduces to efficient nearest-neighbor search over the dense index.

\subsection{Group Relative Policy Optimization}
\label{sec:preliminaries:grpo}

Group Relative Policy Optimization (GRPO)~\cite{guo2025deepseek} optimizes a policy using rewards normalized within a group of sampled candidates, avoiding a separately learned value function. For a group $\mathcal{G}_i$ of $k$ candidates with rewards $r_{ij}$, GRPO computes the group-relative advantage
\begin{align}
    A_{ij}
    &= \frac{r_{ij}-\mu_i}{\sigma_i+\epsilon},
    \label{eq:group_advantage}\\
    \mu_i
    &= \frac{1}{k}\sum_{j\in\mathcal{G}_i}r_{ij},
    \qquad
    \sigma_i^2
    = \frac{1}{k}\sum_{j\in\mathcal{G}_i}(r_{ij}-\mu_i)^2.
    \nonumber
\end{align}
When $\sigma_i=0$, all candidates carry the same reward and therefore provide no relative ranking signal; we set their advantages to zero.

Let $\pi_\theta$, $\pi_{\mathrm{old}}$, and $\pi_{\mathrm{ref}}$ denote the current, rollout, and reference policies, respectively. The clipped GRPO objective with reference-policy regularization is
\begin{align}
    \mathcal{L}_{\mathrm{GRPO}}(\theta)
    ={}&-\mathbb{E}_{i}\!\left[
       \frac{1}{k}\sum_{j\in\mathcal{G}_i}
       \min\!\left(
           \rho_{ij} A_{ij},
           \operatorname{clip}(\rho_{ij},1-\varepsilon,1+\varepsilon)
           A_{ij}
       \right)\right]
       \nonumber\\
       &+\beta D_{\mathrm{KL}}(\pi_\theta\|\pi_{\mathrm{ref}}),
    \label{eq:grpo_loss}
\end{align}
where $\rho_{ij}=\pi_\theta(j)/\pi_{\mathrm{old}}(j)$ is the importance ratio, $\varepsilon$ is the clipping threshold, and $\beta$ controls an optional KL penalty toward the reference policy. TraVEL instantiates the policy, candidate groups, and rewards for driving-video retrieval in Sec.~\ref{sec:method:grpo}.

%% file: sec/4_methodology.tex
\section{TraVEL: Trajectory-Guided Video Embedding Learning}
\label{sec:methodology}

The single-vector retrieval formulation in Sec.~\ref{sec:preliminaries:retrieval} keeps search efficient, but does not by itself ensure that the embedding space preserves fine-grained vehicle motion. During training, TraVEL augments each text--video pair $(q_i,x_i)$ with a synchronized ego-pose sequence $\mathbf{P}_i$. These poses provide privileged training supervision only: they are never provided as inputs to the embedding model and are not required when indexing or retrieving videos. In the following sections, we first diagnose the motion-related failure modes of a general-purpose embedding model, adapt it to the driving domain with paired caption supervision, and finally reshape its video representation using a trajectory-derived GRPO objective.

\subsection{Diagnosing Driving-Video Embedding Gaps}
\label{sec:method:analysis}

\begin{figure*}[t]
    \centering
    \begin{minipage}[t]{0.48\linewidth}
        \centering
        \includegraphics[width=\linewidth]{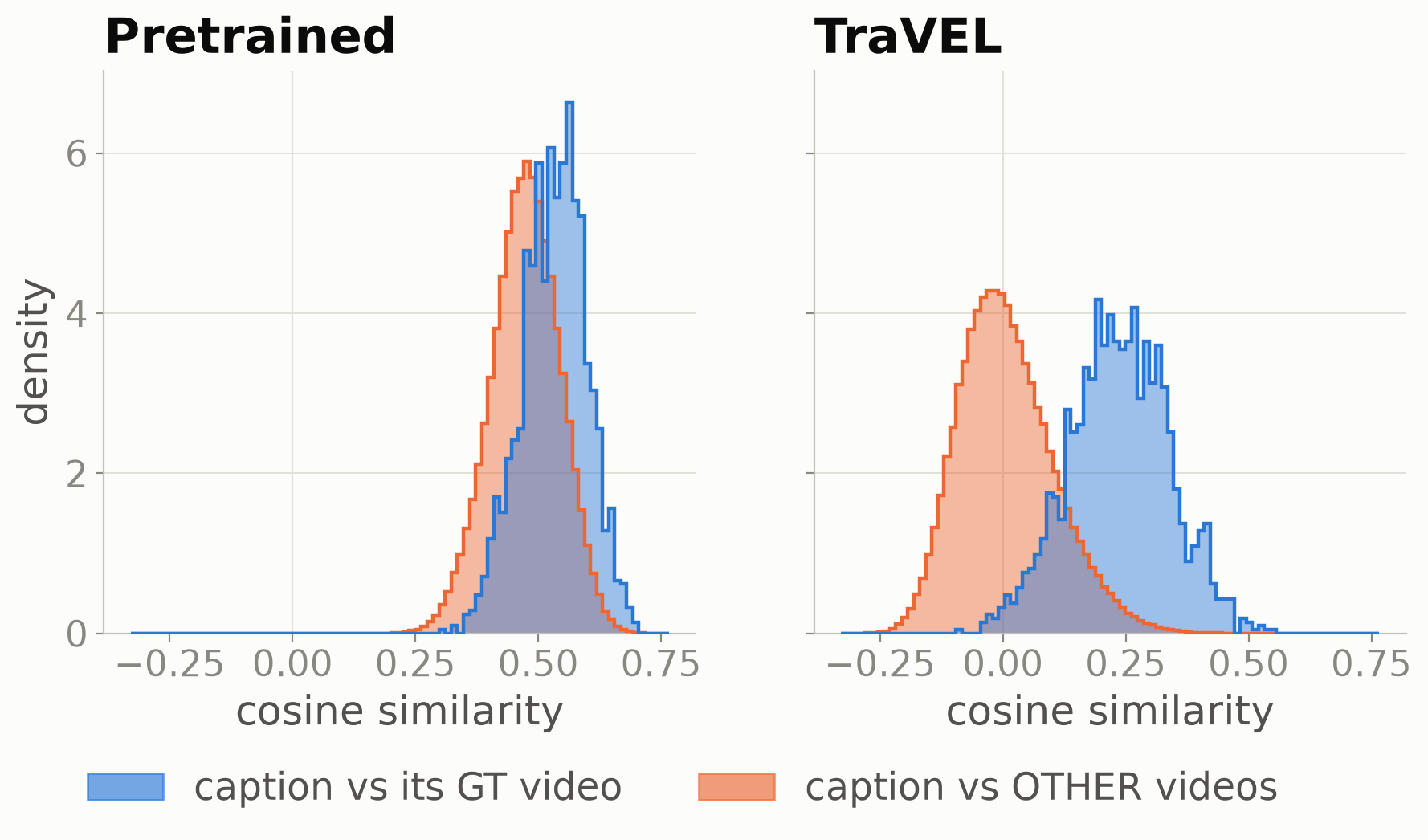}
        \par\smallskip
        {\small \textbf{(a)} Caption--video similarity distributions.}
    \end{minipage}\hfill%
    \begin{minipage}[t]{0.48\linewidth}
        \centering
        \includegraphics[width=\linewidth]{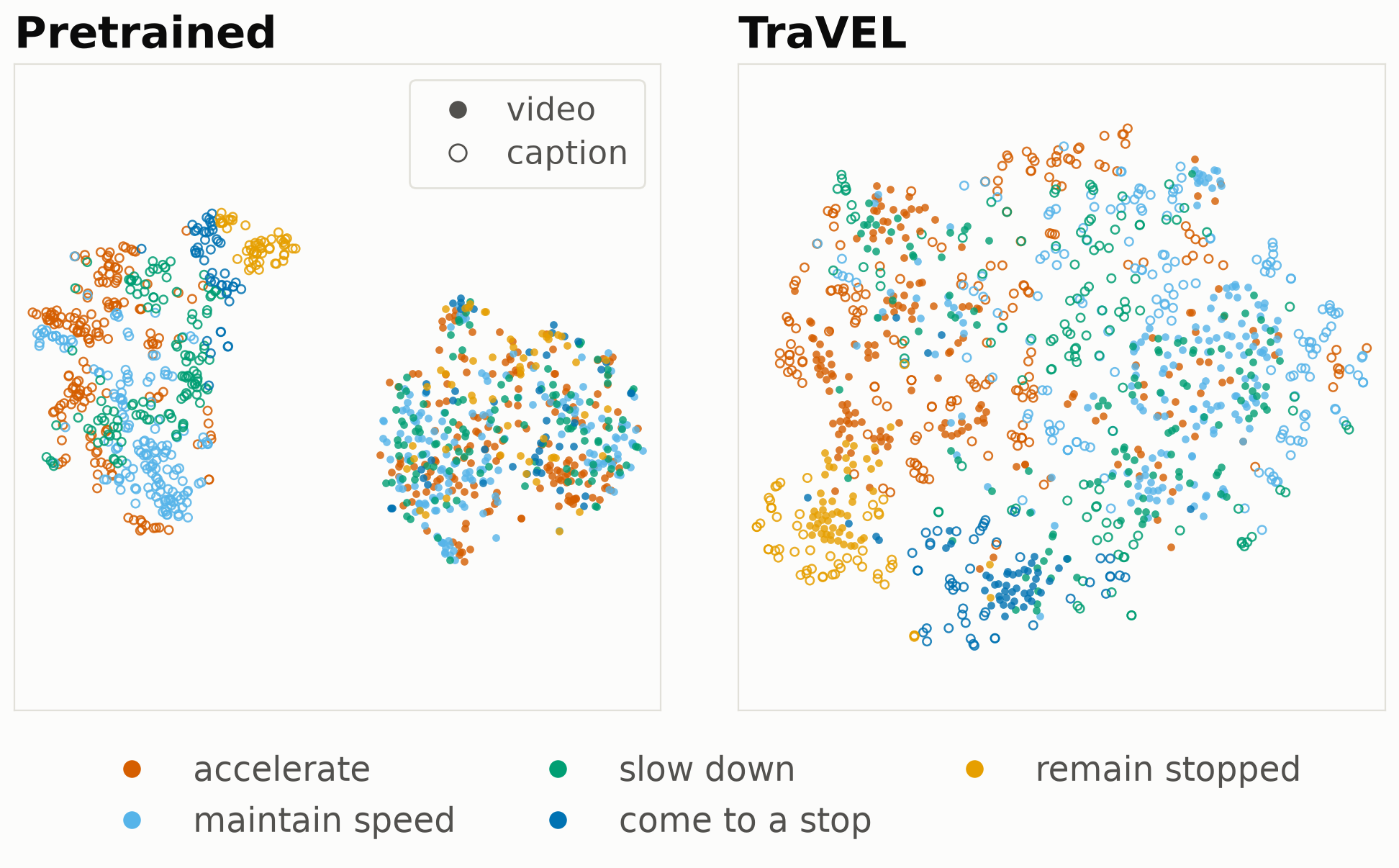}
        \par\smallskip
        {\small \textbf{(b)} Joint caption--video embedding visualization.}
    \end{minipage}
    \caption{Diagnosis of the pretrained driving-video embedding space, with the representation after the complete TraVEL pipeline shown for comparison. (a) Density of cosine similarity between each reasoning-trace caption and its paired ground-truth video (blue) or non-ground-truth videos (orange). Their heavy overlap under Pretrained demonstrates weak retrieval discriminability; the TraVEL panel previews the separation obtained after fine-tuning. (b) Two-dimensional t-SNE projection of video embeddings (filled markers) and caption embeddings (hollow markers), with colors denoting longitudinal motion categories. Pretrained exhibits a large caption--video modality gap and strongly intermixed video motions, while the TraVEL panel shows the resulting cross-modal alignment and clearer motion-specific structure.}
    \label{fig:initial_embedding_analysis}
\end{figure*}

We begin with Qwen3-VL-Embedding~\cite{li2026qwen3vlembedding} and directly evaluate its pretrained representations on driving videos. As shown in Fig.~\ref{fig:initial_embedding_analysis}(a), using the pretrained model for retrieval yields poor performance. The figure compares the cosine-similarity distribution between each reasoning-trace caption and its paired ground-truth video with that between the same caption and all other videos. For the pretrained model, the two distributions are concentrated at similarly high values and overlap substantially. Consequently, paired and non-paired videos receive nearly indistinguishable similarity scores, leaving little discriminative margin for retrieval.

We further examine the structure of the embedding space. Fig.~\ref{fig:initial_embedding_analysis}(b) presents a t-distributed stochastic neighbor embedding (t-SNE) visualization~\cite{vandermaaten2008visualizing} of caption and video embeddings, colored by longitudinal-motion category. A clear modality gap separates caption embeddings from video embeddings, making cross-modal matching difficult. Moreover, video embeddings associated with different motion categories are densely intermixed, indicating that the pretrained representation does not organize driving clips according to ego-vehicle motion.

We attribute these limitations to two factors: the domain gap between general-purpose pretraining data and driving videos, and insufficient motion sensitivity because textual supervision provides only an indirect signal for fine-grained vehicle dynamics. To address both issues, we propose a two-stage training pipeline. The first stage uses supervised fine-tuning to adapt the embedding space to the driving domain, while the second stage applies trajectory-aware GRPO to explicitly improve motion sensitivity.

\subsection{Driving-Domain Supervised Fine-Tuning}
\label{sec:method:sft}

We first reduce the visual and linguistic domain gap through supervised contrastive fine-tuning. Given a mini-batch $\mathcal{B}$ of videos paired with reasoning-trace captions, every non-matching video in the batch serves as a negative for the corresponding caption. The text-to-video InfoNCE objective~\cite{oord2018representation} is
\begin{equation}
    \mathcal{L}_{\mathrm{SFT}}(\theta)
    = -\frac{1}{|\mathcal{B}|}
    \sum_{i\in\mathcal{B}}
    \log
    \frac{\exp\!\left(s_\theta(q_i,x_i)/\tau_{\mathrm{SFT}}\right)}
    {\sum_{j\in\mathcal{B}}\exp\!\left(s_\theta(q_i,x_j)/\tau_{\mathrm{SFT}}\right)},
    \label{eq:sft_infonce}
\end{equation}
where $\tau_{\mathrm{SFT}}$ is a temperature. This stage aligns driving-specific language with the corresponding visual content and yields the checkpoint $\theta_{\mathrm{SFT}}$ used to initialize the reinforcement stage.

Caption supervision substantially improves overall retrieval, but it provides only a sparse description of the motion contained in each clip. Moreover, InfoNCE treats every non-matching video as a negative: two clips with nearly identical ego motion receive no more affinity than two physically unrelated clips. This binary supervision is poorly matched to the continuous structure of vehicle dynamics. The remaining errors on motion-centric queries motivate a second stage that uses the full ego trajectory as dense supervision.

\subsection{Trajectory-Aware GRPO Fine-Tuning}
\label{sec:method:grpo}

\begin{figure*}[t]
    \centering
    \includegraphics[width=\textwidth]{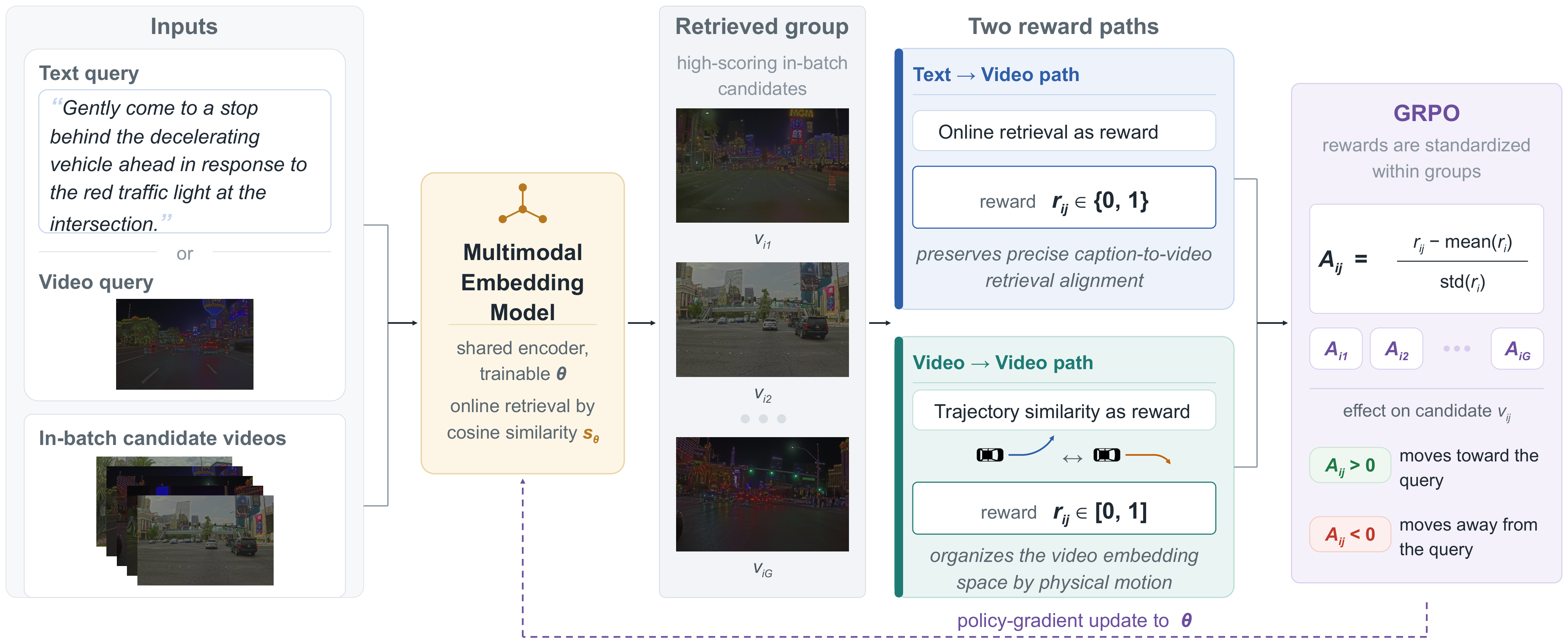}
    \caption{Overview of TraVEL's trajectory-aware GRPO fine-tuning. Given a video or text query and the in-batch video candidates, the current embedding policy performs online retrieval to form a group of high-scoring candidates. The text-to-video group combines the paired video with online hard negatives and uses binary instance rewards, whereas the video-to-video group uses graded ego-trajectory similarity rewards. GRPO standardizes rewards within each group and updates the shared embedding model, preserving instance-level retrieval while organizing the video space by motion.}
    \label{fig:grpo_architecture}
\end{figure*}

Our reinforcement stage uses Group Relative Policy Optimization (GRPO)~\cite{guo2025deepseek} and physical ego motion to organize the video embedding space while retaining a text-to-video retrieval objective. Fig.~\ref{fig:grpo_architecture} provides an overview of this training stage.

\paragraph{Motion representation.}
Each clip spans the same temporal window as an ego-pose sequence, resampled to $T$ steps. From this sequence, we extract scalar motion channels
\begin{equation}
    \mathbf{u}_i^c
    = \big[u_i^c(1),\ldots,u_i^c(T)\big],
    \qquad c\in\mathcal{C}.
    \label{eq:motion_channels}
\end{equation}
Our implementation uses ego-frame lateral displacement and speed. We keep the two channels separate because their physical units and numerical ranges differ; directly concatenating them would allow the largest-scale channel to dominate the reward.
To better capture behavior changes in the lower-value region, such as the transition from stopping to accelerating, we apply the following warping function to each signal:\begin{equation}g_c(u)= \operatorname{sign}(u)\log!\left(1+\frac{|u|}{u_{0,c}}\right),\label{eq}\end{equation}where $u_{0,c}>0$ is a channel-specific reference scale. The transformation preserves the signal direction while compressing large values, thereby placing greater emphasis on changes near zero during temporal alignment.

To compare two motion signals, we use dynamic time warping (DTW)~\cite{sakoe1978dynamic}, which aligns similar maneuvers even when they occur at slightly different times within a clip. For clips $i$ and $j$, the DTW distance for channel $c$ is
\begin{align}
    d_c(i,j)
    &= \frac{1}{T}\,
       \mathrm{DTW}\!\left(g_c(\mathbf{u}_i^c),g_c(\mathbf{u}_j^c)\right),
       \label{eq:channel_dtw}\\
    \mathrm{DTW}(\mathbf{a},\mathbf{b})
    &= \min_{\gamma\in\Gamma}
       \sum_{(t,t')\in\gamma}|a(t)-b(t')|,
    \nonumber
\end{align}
where $\Gamma$ denotes the set of monotone alignment paths. The pointwise cost captures differences in motion direction and magnitude, while DTW provides tolerance to temporal offsets.

Because motion channels may have different numerical scales, we normalize each channel using a robust corpus-level scale $D_c$, defined as the $q$-th percentile of its pairwise distances. The normalized distance is $\widehat{d}_c(i,j)=\min(d_c(i,j)/D_c,1)$, where clipping prevents extreme trajectories from dominating the reward. We then combine the channels using non-negative weights $w_c$:
\begin{equation}
    r_{ij}^{\mathrm{mot}}
    = R\left(
        1-
        \frac{\sum_{c\in\mathcal{C}}w_c\widehat{d}_c(i,j)}
             {\sum_{c\in\mathcal{C}}w_c}
      \right).
    \label{eq:trajectory_reward}
\end{equation}
The resulting reward lies in $[0,R]$, with larger values assigned to clips exhibiting more similar ego motion. Normalizing by $\sum_c w_c$ keeps the reward range unchanged when the channel weights are adjusted.

\paragraph{On-policy retrieval groups.}
At each optimization step, we construct candidate groups from the current model scores over a candidate pool $\mathcal{P}$. For each paired caption--video example $(q_i,x_i)$, we define
\begin{align}
    \mathcal{G}^{\mathrm{t2v}}_i
    &= \{i\}\cup
       \operatorname{Top}_{k-1}
       \big\{s_\theta(q_i,x_j):j\in\mathcal{P},j\neq i\big\},
       \label{eq:t2v_group}\\
    \mathcal{G}^{\mathrm{v2v}}_i
    &= \operatorname{Top}_{k}
       \big\{s_\theta(x_i,x_j):j\in\mathcal{P},j\neq i\big\}.
       \label{eq:v2v_group}
\end{align}
The text-to-video group contains the paired video and the highest-scoring non-paired videos under the current model. We assign the binary reward
$r_{ij}^{\mathrm{t2v}}=\mathbf{1}[j=i]$, encouraging the model to preserve the paired text--video alignment learned during supervised fine-tuning. The video-to-video group excludes the query video itself and instead uses the trajectory-based reward from Eq.~\eqref{eq:trajectory_reward}, such that
$r_{ij}^{\mathrm{v2v}}=r_{ij}^{\mathrm{mot}}$.

We apply the graded trajectory reward to video-to-video similarity rather than directly to text-to-video retrieval. A motion description can correspond to multiple valid clips, and imposing this many-to-many relationship on the text-to-video pathway may conflict with paired-caption supervision. Instead, the video-to-video objective organizes the video embedding space according to ego motion, while the text-to-video objective maintains precise cross-modal alignment.

\paragraph{Group-relative optimization.}
For modality $m\in\{\mathrm{t2v},\mathrm{v2v}\}$, let
$a_i^{\mathrm{t2v}}=q_i$ and $a_i^{\mathrm{v2v}}=x_i$. We convert the retrieval scores within each group into a categorical policy:
\begin{equation}
    \pi_\theta^m(j\mid a_i^m)
    = \frac{\exp\!\left(s_\theta(a_i^m,x_j)/\tau\right)}
           {\sum_{j'\in\mathcal{G}_i^m}
            \exp\!\left(s_\theta(a_i^m,x_{j'})/\tau\right)},
    \qquad j\in\mathcal{G}_i^m.
    \label{eq:retrieval_policy}
\end{equation}
For each group, rewards are standardized using Eq.~\eqref{eq:group_advantage}, and the resulting advantages are optimized with the GRPO objective in Eq.~\eqref{eq:grpo_loss}. This yields separate text-to-video and video-to-video objectives, denoted by $\mathcal{L}^{\mathrm{t2v}}$ and $\mathcal{L}^{\mathrm{v2v}}$, respectively. The final objective combines cross-modal retrieval alignment with motion-aware video-space learning:
\begin{equation}
    \mathcal{L}_{\mathrm{GRPO}}(\theta)
    = \mathcal{L}^{\mathrm{t2v}}(\theta)
      + \lambda\mathcal{L}^{\mathrm{v2v}}(\theta).
    \label{eq:full_grpo_objective}
\end{equation}

%% file: sec/5_experiments.tex
\section{Experiments}
\label{sec:experiments}

\subsection{Experimental Setup}
\label{sec:experimental_setup}

\paragraph{Dataset.}
nuReasoning~\cite{huang2026nureasoning} contains 20{,}000 long-tail driving clips, each spanning 20 seconds. At the time of our experiments, the released subset contains 2{,}589 source clips, which we partition into disjoint training and test sets. From each 20-second source clip, we extract three 8-second windows anchored at 5, 10, and 15 seconds. Each window contains 3 seconds of history and 5 seconds of future context relative to its anchor. We use the nuReasoning reasoning trace associated with each clip as its natural-language caption for supervised fine-tuning and instance-level retrieval. 
Synchronized ego poses are used to construct the trajectory reward during TraVEL's GRPO stage.

\paragraph{Evaluation metrics.}
We evaluate both instance-level retrieval and motion-centric retrieval. For instance-level retrieval, each reasoning-trace caption is used to rank all 1{,}715 videos in the held-out pool by cosine similarity. We report recall at rank $K$ (R@$K$), the percentage of queries whose paired video appears among the top $K$ results, for $K\in\{1,5,10\}$. We additionally report the median rank (MdR) and mean rank (MnR) of the paired video across queries. Higher R@$K$ and lower MdR and MnR indicate better retrieval. To measure motion understanding independently of exact caption matching, we additionally use each longitudinal or lateral motion phrase as a text query and rank the same video pool using its similarity to the video embeddings. We compute average precision (AP) against the videos associated with that motion and report mean average precision (mAP) by averaging AP across all phrases in each motion family. R@$K$, AP, and mAP are reported as percentages, whereas MdR and MnR are raw rank positions.

\paragraph{Implementation details.}
We compare against four off-the-shelf video--text embedding models spanning a range of architectures and scales: CLIP4Clip~\cite{luo2021clip4clip} with a ViT-B/32 backbone and mean pooling, InternVideo2-Stage2~\cite{wang2024internvideo2}, the 224p variant of Cosmos-Embed1~\cite{nvidia2025cosmosembed1}, and Qwen3-VL-Embed~\cite{li2026qwen3vlembedding}. We evaluate their publicly released pretrained checkpoints without adaptation to nuReasoning and use each model's native preprocessing. In all cases, the reasoning trace and its paired 8-second clip are encoded independently, and candidate videos are ranked by cosine similarity between the resulting query and video embeddings.

To study adaptation at different model scales, we use the 2B and 8B Qwen3-VL-Embed checkpoints. We refer to the original checkpoints as \emph{Pretrained}. \emph{SFT} applies the contrastive objective in Eq.~\eqref{eq:sft_infonce}, while \emph{TraVEL} combines this supervised fine-tuning stage with trajectory-aware GRPO. We follow the VLM2Vec recipe~\cite{jiang2024vlm2vec} for multimodal preprocessing and supervised fine-tuning. Tables~\ref{tab:overall_retrieval} and~\ref{tab:motion_retrieval} report instance-level and motion-centric retrieval performance, respectively, at both model scales.

For trajectory-aware GRPO, each 8-second clip is represented by $T=80$ poses sampled at 10\,Hz. We apply logarithmic warping with reference scales of $1$\,m for lateral displacement and $1$\,m/s for speed. Using the 95th percentile for normalization gives channel scales of $D_c=1.264$ and $2.342$, respectively. The lateral-displacement and speed channels are weighted by $1{:}2$, and the maximum trajectory reward is $R=10$. Each optimization step uses a global candidate pool of 768 examples, from which we construct text-to-video and video-to-video groups with $k=24$ candidates each. We set $\tau=0.05$, $\varepsilon=0.2$, $\beta=0$, and $\lambda=1$. The trajectory reward is computed only for the selected video pairs after group construction, keeping the DTW overhead small relative to the encoder forward pass and avoiding the need for a corpus-scale reward table.

\subsection{Results}
\label{sec:experimental_results}

\begin{table*}[t]
    \centering
    \caption{Instance-level text-to-video retrieval on the held-out nuReasoning pool. Recall is reported as a percentage, while MdR and MnR are raw rank positions. The chance-level reference is the analytical expectation under a uniformly random ranking of the 1{,}715 candidate videos. The best result in each column is bold.}
    \label{tab:overall_retrieval}
    \begin{tabular}{@{}lcccccc@{}}
        \toprule
        Model & \#Params & R@1 $\uparrow$ & R@5 $\uparrow$ & R@10 $\uparrow$ & MdR $\downarrow$ & MnR $\downarrow$ \\
        \midrule
        \multicolumn{7}{@{}l}{\textit{Chance-level reference}} \\
        Uniform random ranking & -- & 0.1 & 0.3 & 0.6 & 858 & 858.0 \\
        \midrule
        \multicolumn{7}{@{}l}{\textit{Off-the-shelf models}} \\
        CLIP4Clip & 151M & 0.6 & 2.6 & 4.7 & 415 & 545.0 \\
        InternVideo2 & 6B & 0.1 & 0.6 & 1.0 & 860 & 858.6 \\
        Cosmos-Embed1 & 1.2B & 2.4 & 9.2 & 14.8 & 106 & 231.7 \\
        Qwen3-VL-Embed & 2B & 1.3 & 4.5 & 7.4 & 266 & 400.5 \\
        Qwen3-VL-Embed & 8B & 1.6 & 5.2 & 7.8 & 264 & 371.7 \\
        \midrule
        \multicolumn{7}{@{}l}{\textit{Fine-tuned models}} \\
        Qwen3-VL-Embed + TraVEL & 2B & 8.7 & 24.5 & 35.3 & 23 & 87.8 \\
        Qwen3-VL-Embed + TraVEL & 8B & \textbf{10.1} & \textbf{28.3} & \textbf{39.8} & \textbf{17} & \textbf{70.6} \\
        \bottomrule
    \end{tabular}
\end{table*}

\begin{table*}[t]
    \centering
    \caption{Motion-centric text-to-video retrieval at two model scales. Each motion phrase is embedded as a text query and evaluated against the held-out video pool. Per-query AP is reported for individual motions, and mAP averages AP within each motion family. All values are percentages; higher is better. The best result within each model scale is bold.}
    \label{tab:motion_retrieval}
    \small
    \setlength{\tabcolsep}{5pt}
    \begin{tabular}{@{}lcccccc@{}}
        \toprule
        \multirow{2}{*}{Motion query} & \multicolumn{3}{c}{2B} & \multicolumn{3}{c}{8B} \\
        \cmidrule(lr){2-4}\cmidrule(lr){5-7}
        & Pretrained & SFT & TraVEL & Pretrained & SFT & TraVEL \\
        \midrule
        \multicolumn{7}{@{}l}{\textit{Longitudinal motion}} \\
        Gently accelerate & 34.1 & 49.9 & \textbf{55.7} & 29.2 & 49.9 & \textbf{56.6} \\
        Maintain speed & 39.0 & 44.2 & \textbf{49.2} & 30.0 & 44.1 & \textbf{49.7} \\
        Slow down gently & 25.2 & \textbf{30.7} & 30.4 & 32.3 & 32.0 & \textbf{34.2} \\
        Gently come to a stop & 14.7 & 33.7 & \textbf{53.8} & 18.4 & 43.3 & \textbf{57.4} \\
        Remain stopped & 10.6 & 70.8 & \textbf{89.4} & 21.8 & 82.0 & \textbf{89.4} \\
        \cmidrule(lr){1-7}
        \textit{mAP} & 24.7 & 45.9 & \textbf{55.7} & 26.3 & 50.3 & \textbf{57.5} \\
        \midrule
        \multicolumn{7}{@{}l}{\textit{Lateral motion}} \\
        No lateral action & 66.2 & 77.8 & \textbf{80.5} & 63.4 & 77.1 & \textbf{78.9} \\
        Turn right & 13.1 & 56.0 & \textbf{69.8} & 19.1 & 70.5 & \textbf{72.0} \\
        Turn left & 16.5 & 49.7 & \textbf{63.6} & 14.4 & 65.6 & \textbf{68.3} \\
        Right lane change & 9.8 & 9.0 & \textbf{10.2} & 10.5 & 19.4 & \textbf{20.2} \\
        Left lane change & 5.7 & 11.1 & \textbf{11.7} & 7.6 & 19.3 & \textbf{22.5} \\
        Slightly move right in the lane & 4.6 & 4.2 & \textbf{4.7} & \textbf{7.7} & 6.4 & 6.8 \\
        Slightly move left in the lane & 8.1 & 9.1 & \textbf{9.6} & 4.9 & 10.2 & \textbf{10.6} \\
        \cmidrule(lr){1-7}
        \textit{mAP} & 17.7 & 31.0 & \textbf{35.7} & 18.2 & 38.4 & \textbf{39.9} \\
        \bottomrule
    \end{tabular}
\end{table*}

\paragraph{Overall retrieval.}
As shown in Tab.~\ref{tab:overall_retrieval}, all off-the-shelf models exhibit a substantial domain gap on driving-video retrieval: their R@1 values are at most 2.4, despite ranging from 151M to 8B parameters. InternVideo2 remains close to the uniform-random reference, whereas Cosmos-Embed1 is the strongest off-the-shelf baseline, reaching 2.4, 9.2, and 14.8 at R@1, R@5, and R@10, respectively, with a median rank of 106. Scaling Qwen3-VL-Embed from 2B to 8B yields only marginal zero-shot gains, increasing R@1 from 1.3 to 1.6 and R@10 from 7.4 to 7.8. These results indicate that model scale alone does not resolve the driving-domain gap.

Fine-tuning with TraVEL substantially improves instance retrieval at both model scales. At 2B, TraVEL increases R@1, R@5, and R@10 from 1.3, 4.5, and 7.4 to 8.7, 24.5, and 35.3, respectively, while reducing the median rank from 266 to 23. At 8B, it raises the corresponding recall values from 1.6, 5.2, and 7.8 to 10.1, 28.3, and 39.8, and reduces the median rank from 264 to 17. These consistent gains show that the TraVEL pipeline effectively adapts a general-purpose embedding model to driving-video retrieval.

\paragraph{Motion-centric retrieval.}
Table~\ref{tab:motion_retrieval} separates the effects of supervised fine-tuning and trajectory-aware optimization. At 2B, SFT improves longitudinal and lateral mAP from 24.7 and 17.7 to 45.9 and 31.0, respectively. TraVEL further raises them to 55.7 and 35.7, corresponding to gains of 9.8 and 4.7 points over SFT. The same trend holds at 8B: SFT reaches 50.3 longitudinal and 38.4 lateral mAP, while TraVEL improves them to 57.5 and 39.9, for additional gains of 7.2 and 1.5 points. The largest improvements appear in stopping and turning behaviors, indicating that trajectory supervision provides motion structure beyond what is learned from captions alone. Smaller gains on subtle within-lane shifts suggest that these fine-grained lateral distinctions remain challenging.

\begin{figure*}[t]
    \centering
    \includegraphics[width=\textwidth]{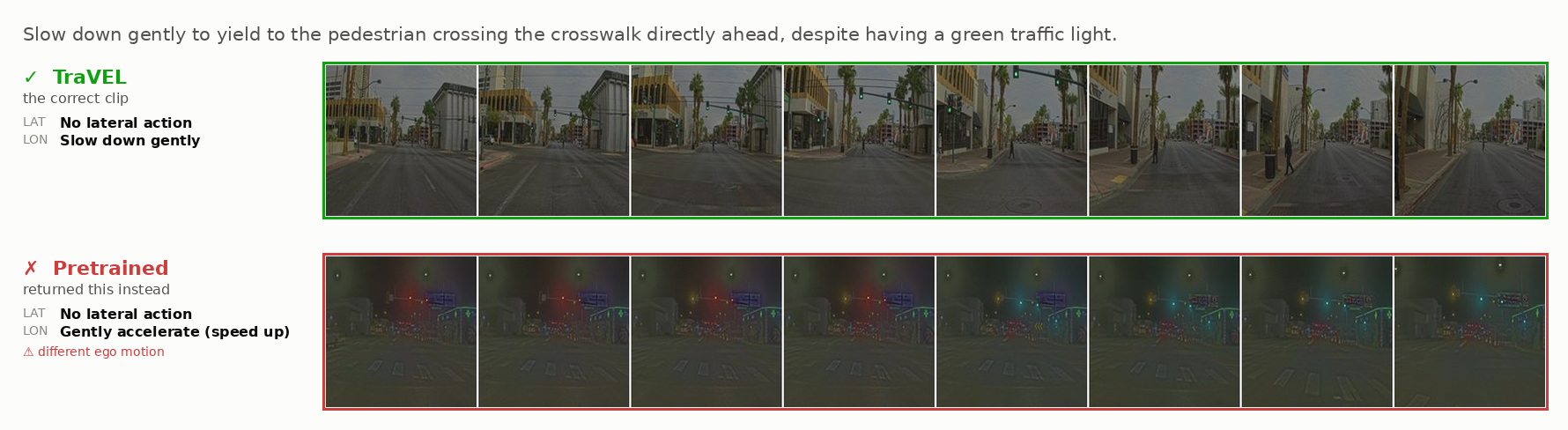}
    \caption{Qualitative top-1 text-to-video retrieval for a nuReasoning reasoning-trace query describing gentle deceleration to yield to a pedestrian despite a green traffic light. TraVEL retrieves the paired ground-truth clip (green), matching both the slowing behavior and pedestrian interaction, whereas Pretrained retrieves an incorrect nighttime clip in which the ego vehicle accelerates (red).}
    \label{fig:qualitative_reasoning_retrieval}
\end{figure*}

\paragraph{Qualitative retrieval.}
Fig.~\ref{fig:qualitative_reasoning_retrieval} contrasts Pretrained and TraVEL on a compositional reasoning-trace query. Pretrained fails to recover the requested longitudinal behavior and returns a nighttime clip in which the ego vehicle accelerates. TraVEL instead ranks the paired clip first, capturing both the gentle slowdown and the pedestrian interaction described by the query. This example suggests that trajectory-aware optimization can improve motion sensitivity while retaining contextual alignment with the reasoning trace.

%% file: sec/6_conclusion.tex
\section{Conclusion}
\label{sec:conclusion}

We presented TraVEL, a simple and efficient framework for adapting a general-purpose multimodal embedding model to driving-video retrieval. Across 2B and 8B Qwen3-VL-Embedding checkpoints, supervised fine-tuning on paired clips and reasoning-trace captions first bridges the driving-domain gap. We then use ego-trajectory similarity as a training-only reward within GRPO to organize the embedding space according to vehicle motion, without introducing additional perception modules or changing the single-vector cosine-similarity retrieval pipeline at inference time. TraVEL preserves the instance-level gains of SFT at both model scales. At 2B, it improves longitudinal mAP from 45.9 to 55.7 and lateral mAP from 31.0 to 35.7 over SFT; at 8B, the corresponding improvements are from 50.3 to 57.5 and from 38.4 to 39.9. These results show that readily available physical motion signals can complement language supervision and produce representations that better capture fine-grained driving behavior.

Our current evaluation uses the available subset of nuReasoning and a retrieval pool of 1{,}715 videos. Evaluating on larger pools, additional datasets, and a broader range of driving conditions will be important for establishing scalability and generalization. Fine-grained lane changes and within-lane shifts also remain challenging, suggesting that ego motion alone does not capture every distinction relevant to driving-video retrieval. Future work could incorporate the motion of surrounding agents and scene structure into the reward while retaining the efficiency of embedding-based search.